\documentclass[conference]{IEEEtran}
\IEEEoverridecommandlockouts
\usepackage{cite}
\usepackage{amsmath,amssymb,amsfonts}
\usepackage{algorithmic}
\usepackage{graphicx}
\usepackage{textcomp}
\usepackage{xcolor}
\usepackage{listings}
\usepackage{subcaption}
\usepackage{eurosym}
\def\BibTeX{{\rm B\kern-.05em{\sc i\kern-.025em b}\kern-.08em
    T\kern-.1667em\lower.7ex\hbox{E}\kern-.125emX}}
\begin{document}

\title{An Overview of the Ludii General Game System\\
}

\author{\IEEEauthorblockN{Matthew Stephenson, {\'E}ric Piette, Dennis J.N.J. Soemers and Cameron Browne}
\IEEEauthorblockA{\textit{Department of Data Science and Knowledge Engineering} \\
\textit{Maastricht University}\\
Maastricht, The Netherlands \\
\texttt{\{matthew.stephenson,eric.piette,dennis.soemers,cameron.browne\}@maastrichtuniversity.nl}}
}

\maketitle

\begin{abstract}
The Digital Ludeme Project (DLP) aims to reconstruct and analyse over 1000 traditional strategy games using modern techniques. One of the key aspects of this project is the development of Ludii, a general game system that will be able to model and play the complete range of games required by this project. Such an undertaking will create a wide range of possibilities for new AI challenges. In this paper we describe many of the features of Ludii that can be used. This includes designing and modifying games using the Ludii game description language, creating agents capable of playing these games, and several advantages the system has over prior general game software. 
\end{abstract}


\section{Introduction}

General game research is one of the most popular areas of game-based academia, and is widely viewed as a suitable means of evaluating AI techniques for a variety of applicable domains and problems \cite{gameaibook}. Several general game systems have been developed to assist with this research field, with the main candidates for such work being the General Game Playing (GGP) system \cite{genesereth05}, the General Video Game AI (GVGAI) framework \cite{gvgaioverview}, and the Arcade Learning Environment (ALE) \cite{aleoverview}. We propose Ludii as a new general game system, that can facilitate many novel and exciting areas of research.

Within this demo paper, we describe the key aspects of the Ludii system that make it an appealing alternative to other general game platforms. We cover the language used for describing games in Ludii, which provides a simple and human understandable approach. We also provide details on AI techniques that will be included with Ludii at launch. The first version of Ludii is expected to be released in August 2019, but will continue to be improved and updated with new games, functionality and organised events for many years afterwards.

\section{Ludii System}
Ludii is a new general game system currently under development~\cite{Piette19}. 
It is based on similar principles to the earlier Ludi system~\cite{browne09}, but uses significantly different mechanisms to achieve much greater generality, extensibility and efficiency. 

Ludii is being designed and implemented primarily to provide answers to the questions raised by the Digital Ludeme Project \cite{ludii1}, but will stand alone as a platform for general games research in the areas of agent development, content generation, game design, player analysis, and education.
Ludii provides many advantages over existing GGP systems, including performance improvements, simpler and clearer game descriptions, as well as a highly evolvable language.

Ludii uses a class grammar approach which automatically generates the game description language from the class hierarchy of its underlying source code \cite{BrowneB16}. This ensures that there is a 1:1 mapping between the Ludii source code and the game description grammar, as any games which are expressed using this grammar are automatically instantiated back into the corresponding library code for compilation.
Ludii can theoretically support any rule, equipment or behaviour that can be programmed in Java. The exact implementation details for each of these definitions are encapsulated within our simplified grammar, which summarises the code to be called. Additional details and examples on the Ludii description language are provided in the next section.


\begin{figure*}
\centering

  \begin{subfigure}{0.03\textwidth}
  ~
  \end{subfigure}
    \begin{subfigure}{0.28\textwidth}
    \includegraphics[width=\textwidth]{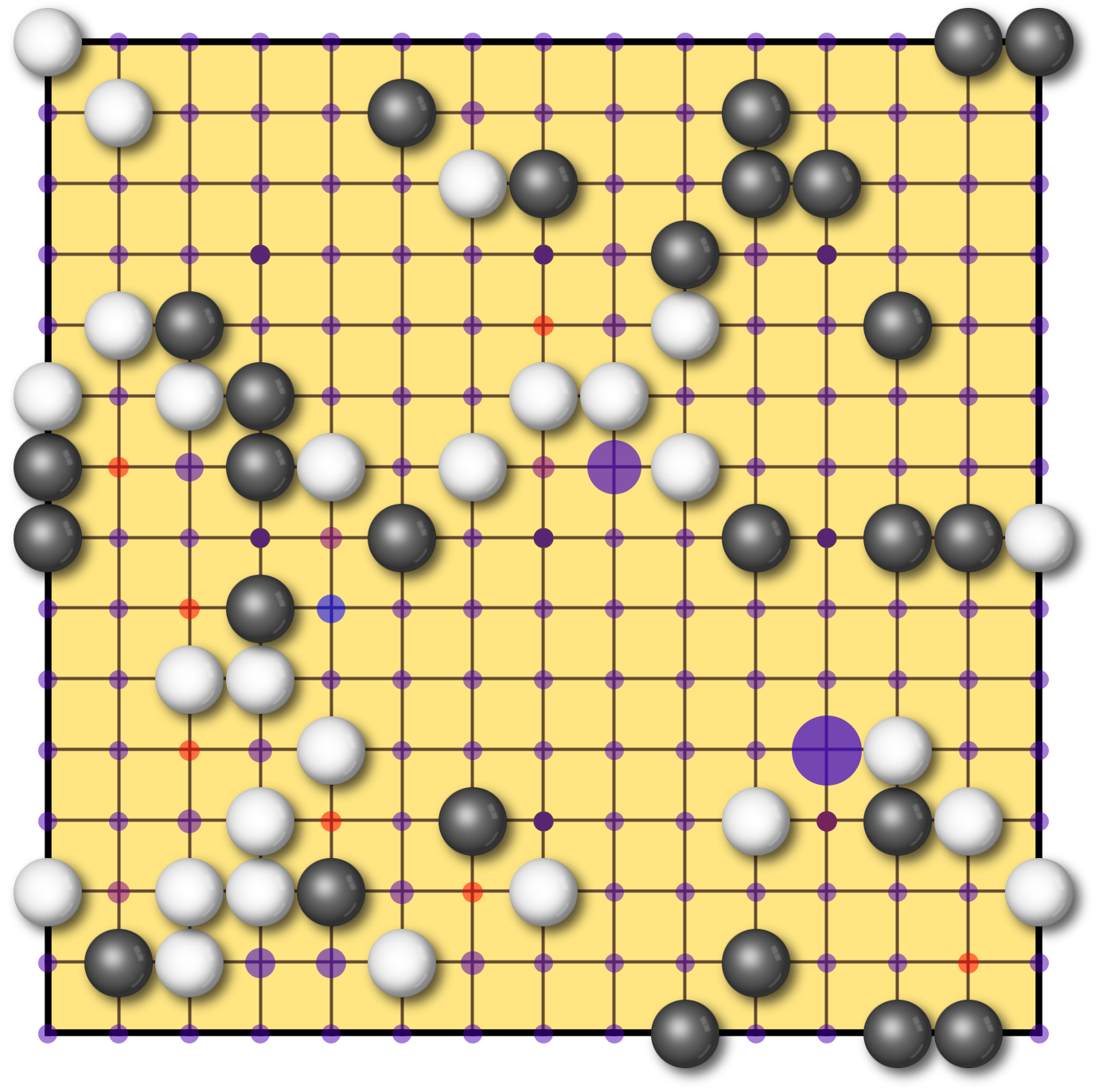}
    \label{fig:1}
  \end{subfigure}
  \begin{subfigure}{0.05\textwidth}
  ~
  \end{subfigure}
  \begin{subfigure}{0.4\textwidth}
  \vspace{-8pt}

\footnotesize
\lstset{basicstyle=\ttfamily}
\begin{lstlisting}
(game "Gomoku"  
  (mode 2)  
  (equipment { 
    (goBoard 15) 
    (ball Each) 
  })  
  (rules 
    (play (to (mover) (empty))) 
    (end (line length:5) (result (mover) Win))  
  )
)
\end{lstlisting}
  \end{subfigure}
  \caption{A game of Gomoku in progress, played out on the Ludii system, along with its ludeme-based game description.}
  \label{Fig:Gomoku}
\end{figure*}

\begin{figure*}[t!]
    \centering
    \begin{subfigure}[t]{0.155\linewidth}
        \centering
        \includegraphics[width=\linewidth]{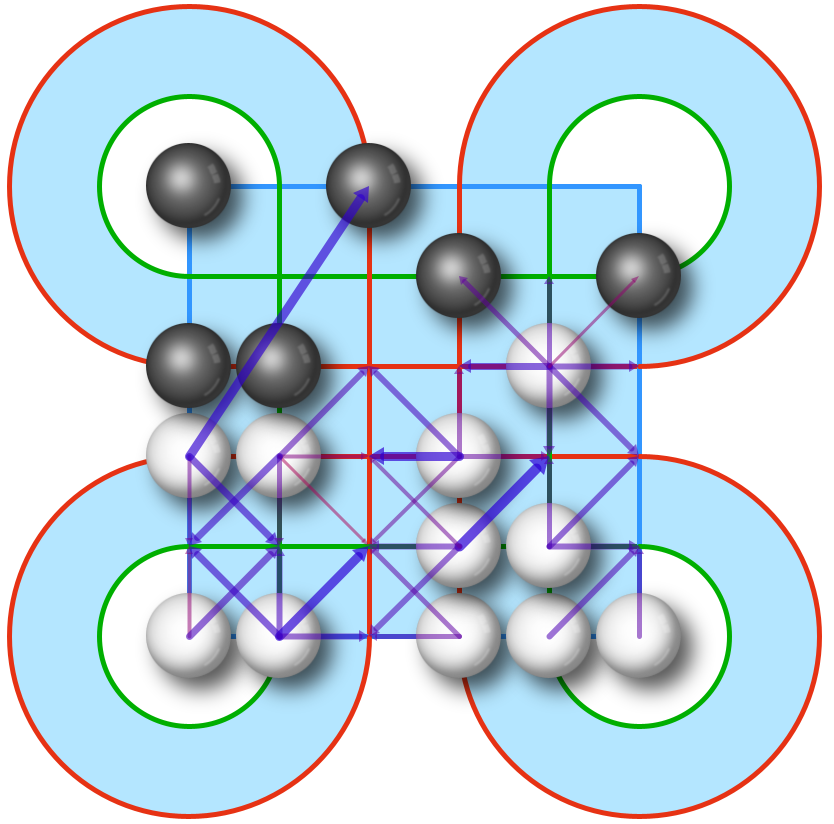}
        \caption{Surakarta}
    \end{subfigure}%
    ~ 
    \begin{subfigure}[t]{0.155\linewidth}
        \centering
        \includegraphics[width=\linewidth]{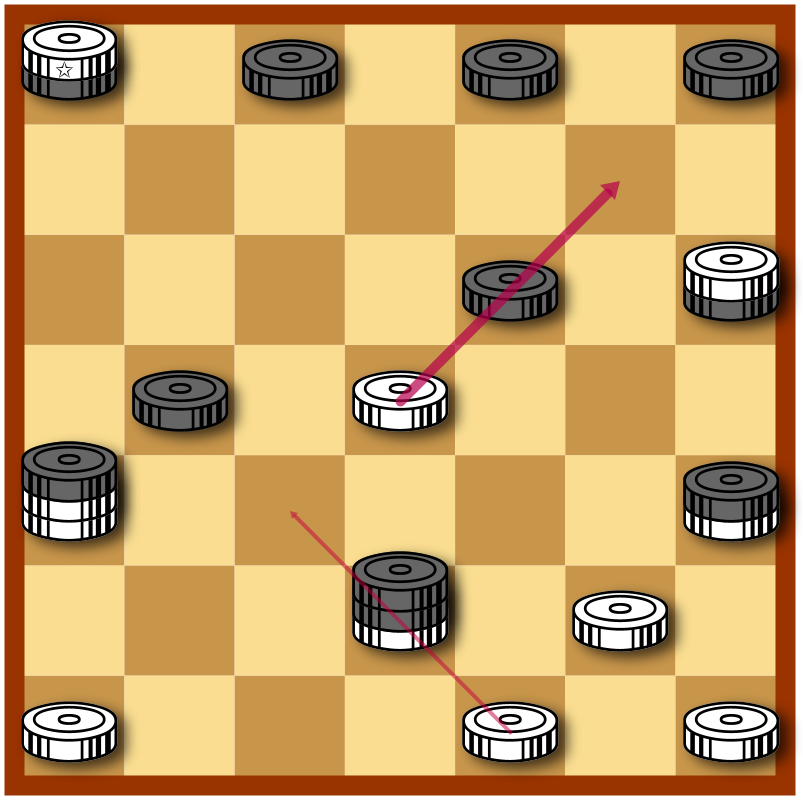}
        \caption{Lasca}
    \end{subfigure}
     ~ 
    \begin{subfigure}[t]{0.135\linewidth}
        \centering
        \includegraphics[width=\linewidth]{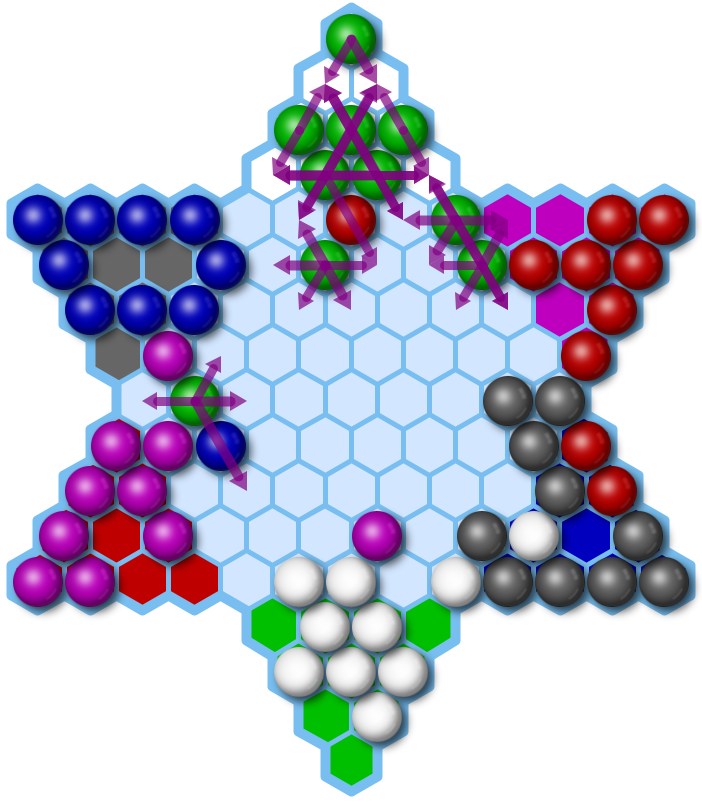}
        \caption{Sternhalma}
    \end{subfigure}
         ~ 
    \begin{subfigure}[t]{0.155\linewidth}
        \centering
        \includegraphics[width=\linewidth]{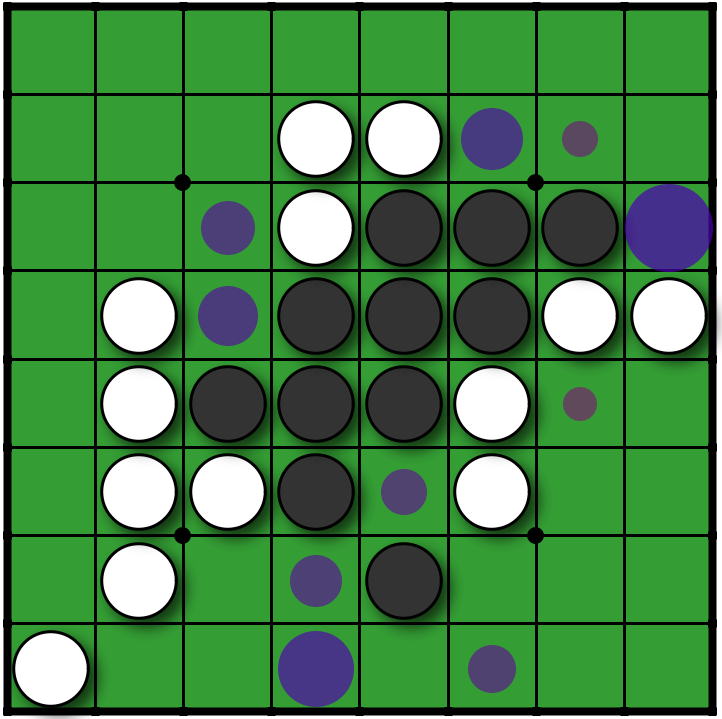}
        \caption{Reversi}
    \end{subfigure}
         ~ 
    \begin{subfigure}[t]{0.155\linewidth}
        \centering
        \includegraphics[width=\linewidth]{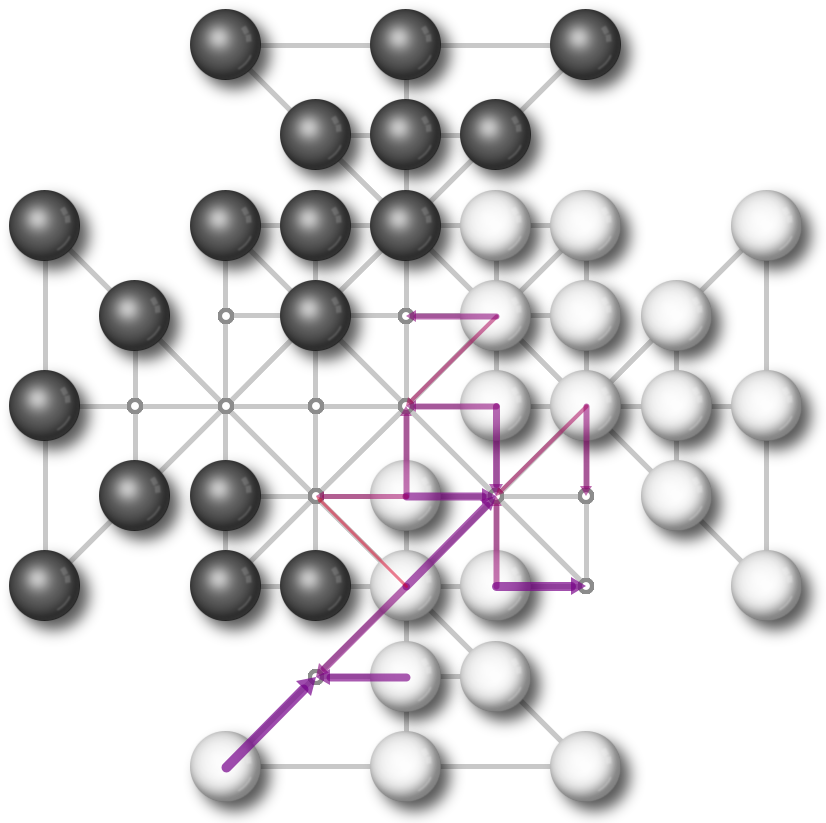}
        \caption{Peralikatuma}
    \end{subfigure}
         ~ 
    \begin{subfigure}[t]{0.155\linewidth}
        \centering
        \includegraphics[width=\linewidth]{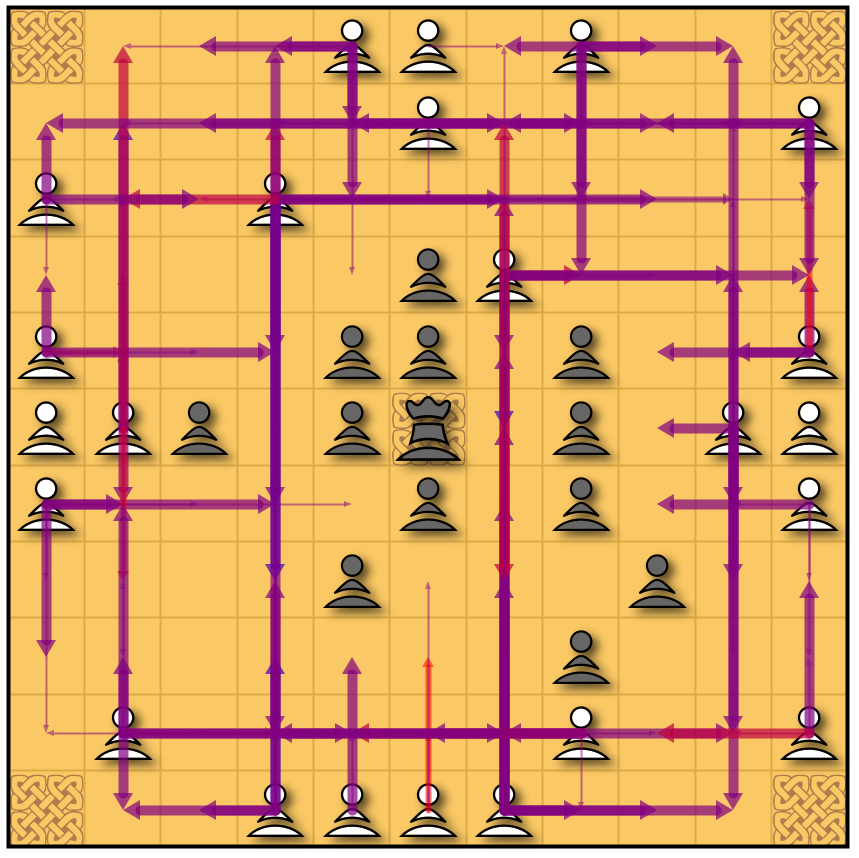}
        \caption{Hnefatafl}
    \end{subfigure}
    \caption{Example agent visualisations for games implemented in Ludii.}
    \label{Fig:ManyGames}
\end{figure*}

A wide range of different game types can be implemented in Ludii, including but not limited to:
\begin{itemize}
  \item Deterministic, stochastic and hidden information games.
  \item Board, card, dice and tile games.
  \item Boardless games (e.g. Dominoes).
  \item Stacking games (e.g. Lasca).
  \item Simultaneous move games (e.g. Chinook).
  \item Graph games (e.g. Dots and Boxes).
  \item Multi-player / team games and single-player puzzles.
\end{itemize}

\section{Ludii Game Description Langugage}

Games are described in the Ludii grammar using {\it  ludemes}, which are conceptual units of game related information that encapsulate key game design concepts, essentially forming the building blocks or ``DNA'' of games. As an example, Figure~\ref{Fig:Gomoku} shows the game of Gomoku, along with the Ludii game description file that was used to create this game. The equipment section describes the pieces and board that are needed to play. The rules section describes how the game was initially set up (start), how each turn in the game proceeds (play), and under what conditions the game is over (end).

Games described using this approach can be easily modified by adjusting ludemes to suit your needs. Changing the board size or victory conditions can be as simple as editing a single parameter. File sizes for each game are also very small, with each game's description fitting into a QR code. The language is easy to read and understand compared to more verbose alternatives such as GDL. Recent experimental results revealed that our system tends to be faster than other GGP alternatives. 
The ability to break games down into their individual ludemes allows for additional analyses, such as phylogenetic analyses, of the similarities and differences between games.

\section{Ludii Agents}

Implementations of standard game-playing agents are included in Ludii. Due to the large number of (variants of) games that are planned to be included, we focus on techniques that are generally applicable across many games. This means that we focus on techniques such as Monte Carlo tree search (MCTS), which is the most popular approach in prior GGP research. Third parties will also be able to develop their own agents for Ludii.

Ludii will also provide visualisations to gain insight into the ``thinking process'' of algorithms such as MCTS. Examples of such visualisations are depicted in Figures~\ref{Fig:Gomoku} and~\ref{Fig:ManyGames}. Arrows are drawn for every possible movement action (in games like Amazons, Chess, Hnefatafl, etc.), and circles are drawn for placement actions (in games like Go, Hex, Reversi, etc.). The sizes of these drawings scale with the visit counts associated with actions in an MCTS search process, and colours are changed based on the value estimates of MCTS (e.g. blue for winning moves, red for losing moves, purple for moves with neutral value estimates).

\section{Conclusion}
The Ludii system presents a new and intuitive approach to game design and game playing. The scope and completeness of the games available with Ludii can rival that of any prior general game playing system. The software is easy to use for those who are not technically inclined, while also providing the functionality for integrating existing agents and AI techniques. The language used to describe games also opens up many new avenues of research, particularly in the areas of general game generation and analysis.

\section*{Acknowledgment.}

This research is part of the European Research Council-funded Digital Ludeme Project (ERC Consolidator Grant \#771292) run by Cameron Browne at Maastricht University's Department of Data Science and Knowledge Engineering. 

\bibliographystyle{IEEEtran}
\bibliography{References}

\end{document}